\documentclass[runningheads]{llncs}
\usepackage[T1]{fontenc}
\usepackage{graphicx}
\usepackage{caption}
\usepackage{amsmath}
\usepackage{amssymb}
\usepackage{cite}
\usepackage{tabularx}
\usepackage{hyperref}
\usepackage{booktabs}
\usepackage{rotating}
\hypersetup{
  colorlinks=true,
  linkcolor=blue,   
  citecolor=blue,   
  urlcolor=blue     
}
\makeatletter
\newcommand{\titlefootnote}[1]{%
  \begingroup
  \renewcommand{\thefootnote}{}%
  \long\def\@makefntext##1{%
    \noindent\footnotesize ##1%
  }%
  \footnotetext{#1}%
  \endgroup
}
\makeatother
\author{Xinyu Jia$^{1}$, Weidong Guo$^{1}$, Wangyuan Zhao$^{1}$, Yi Guo$^{1}$, Zeju Li$^{1,\dagger}$, Yuanyuan Wang$^{1,\dagger}$} 
\institute{College of Biomedical Engineering, Fudan University, Shanghai, China \\
\email{\{zejuli,yywang\}@fudan.edu.cn} 
}

\begin{document}
%
\title{CIPHER: Causal Intervention Pathways for Healthcare Equity and Robustness}

\titlerunning{CIPHER}
\authorrunning{X. Jia et al.}
%
%
%
%
\maketitle              
\titlefootnote{
\textsuperscript{$\dagger$} Corresponding authors.\\
Code is available at \url{https://github.com/fdu-farm/CIPHER}.
}

\begin{abstract}
Deep learning models for medical diagnosis frequently exhibit substantial performance disparities across sensitive subgroups (e.g., race, sex), even when average accuracy is high.
While generative data augmentation offers a route to mitigate this, existing strategies are suboptimal; they typically address only one or two dependency channels between sensitive attributes and image features.
We formalize the medical image formation process via a structural causal model, revealing that sensitive attributes actually influence image content through four distinct pathways—a structural complexity neglected by prior works.
Based on this insight, we introduce CIPHER (\textbf{C}ausal \textbf{I}ntervention \textbf{P}athways for \textbf{H}ealthcare \textbf{E}quity and \textbf{R}obustness), a framework designed to systematically intervene on all four causal paths.
To achieve this, CIPHER utilizes a diffusion backbone equipped with classifier-free guidance and null-text inversion. This technical design enables the faithful reconstruction of patient-specific anatomy while allowing for the precise, editable synthesis of counterfactuals required to break sensitive dependency chains.
We tested CIPHER using chest X-ray and dermoscopy benchmarks across both standard and shifted data distributions. By employing a multi-pathway intervention strategy, our model reduced worst-group disparities by an average of 35.8$\%$ compared to disease-conditioned synthesis baselines, while also improving total diagnostic accuracy.

\end{abstract}

\section{Introduction and Motivation}

Deep learning has achieved strong performance in medical imaging tasks such as chest X-ray (CXR) interpretation and dermoscopic lesion recognition \cite{irvin2019chexpert,johnson2019mimiccxr,cassidy2022isic_analysis,esteva2017skin_cancer}.
Yet strong average accuracy does not guarantee fairness: clinically meaningful gaps often persist across sensitive subgroups defined by sex, age, race/ethnicity, or proxies such as skin tone, which hinders safe and equitable deployment \cite{glocker2023protected_characteristics_cxr,larrazabal2020gender_imbalance,seyyedkalantari2021underdiagnosis_bias,yang2025vlm_demographic_bias,daneshjou2022derm_ai_disparities,chen2023algorithmic_fairness}.
Sensitive attributes can also be implicitly encoded in medical images and exploited as shortcut cues, producing subgroup-specific errors that may worsen under distribution shift across sites, scanners, and protocols \cite{gichoya2022race_recognition,zech2018pneumonia_generalization,geirhos2020shortcut_learning,oakden_rayner2020hidden_stratification,damour2022underspecification}.

Achieving algorithmic fairness requires a rigorous understanding of exactly how sensitive factors propagate through the model's decision-making pipeline. Currently, the field lacks a comprehensive framework to map these complex dynamics. To bridge this gap and move beyond surface-level heuristics, we introduce a formal causality diagram that explicitly models these interactions (Fig.~\ref{fig:framework}).
Let $S$ denote sensitive attribute, $D$ the disease label, $C$ the data availability or sampling  mechanism, and $Y$ the final observed or generated image.
Under this framework, $S$ influences $Y$ via \textbf{four} distinct routes: population bias ($S{\rightarrow}Y$), manifestation bias ($S{\rightarrow}D{\rightarrow}Y$), prevalence bias ($S{\rightarrow}C{\rightarrow}Y$), and environment bias ($S{\rightarrow}D{\rightarrow}C{\rightarrow}Y$).
Because each pathway induces fundamentally different type of model failure, treating synthetic data augmentation as a black-box remedy offers limited insight. Even if overall fairness metrics improve, it remains entirely unclear which specific dependency was corrected (e.g., data coverage, physical manifestation, or spurious shortcuts), ultimately failing to explain why these gains often collapse under distribution shift.

Despite the complex causal landscape of medical image formation, existing fairness interventions remain largely insufficient because they typically target only a single causal edge—primarily through basic attribute resampling~\cite{jones2024nofairlunch, puyol2021fairness} or rudimentary disease-conditioned synthesis~\cite{kazerouni2023diffusion_survey_med,ho2020ddpm,nichol2021improved_ddpm,rombach2022latent_diffusion, ktena2024generative_fairness_shifts}. While generative tools such as inversion-based editing offer a promising avenue for creating instance-anchored counterfactuals that preserve patient-specific context~\cite{mokady2023null_text_inversion,meng2022sdedit}, prior work lacks a unified, causally grounded framework that integrates global distribution reshaping with local editing~\cite{castro2020causality_matters}. We argue that these existing synthesis strategies are fundamentally limited because they address only one or two of the four critical causal pathways, failing to provide a comprehensive solution to the multifaceted nature of subgroup disparities.

Therefore, we propose CIPHER (\textbf{C}ausal \textbf{I}ntervention \textbf{P}athways for \textbf{H}ealthcare \textbf{E}quity and \textbf{R}obustness), a diffusion-based generative framework grounded in the structural causality of medical image formation \cite{jones2024nofairlunch,castro2020causality_matters}.
Unlike existing heuristics, CIPHER provides a complete theoretical safeguard: by formally organizing control around $(S,D,C,Y)$ and intervening on all relevant edges
($S{\rightarrow}Y$, $D{\rightarrow}Y$, $S{\rightarrow}C$, $D{\rightarrow}C$), it comprehensively blocks every path where $S$ can bias the observed distribution. To implement this, our novel architecture leverages classifier-free guidance and null-text inversion to achieve highly faithful, instance-anchored synthesis~\cite{mokady2023null_text_inversion}. Evaluations on CXR and dermoscopy show that mixing complementary paths maximizes accuracy and fairness, consistently reducing worst-group AUROC gaps across distribution shifts. Summary of contributions:
\begin{itemize}
\item We formalize the causal pathways through which sensitive attributes $S$ bias observed images $Y$, laying a foundation for evaluating fairness interventions.
\item We introduce CIPHER, a versatile diffusion-based framework that translates a theoretical causal graph over $(S,D,C,Y)$ into four explicit, intervenable generation and editing pathways.
\item We enable precise demographic subgroup reshaping and multi-attribute, instance-anchored counterfactual editing via null-text inversion.
\item We demonstrate that complementary paths maximizes fairness and robustness, achieving an 8.5$\%$ to 61.6$\%$ reduction in worst-group AUROC gaps under distribution shift compared to existing approaches.
\end{itemize}
\section{Analysis}
\label{sec:analysis}

\begin{figure}[!t]
  \centering
  \includegraphics[width=1\linewidth]{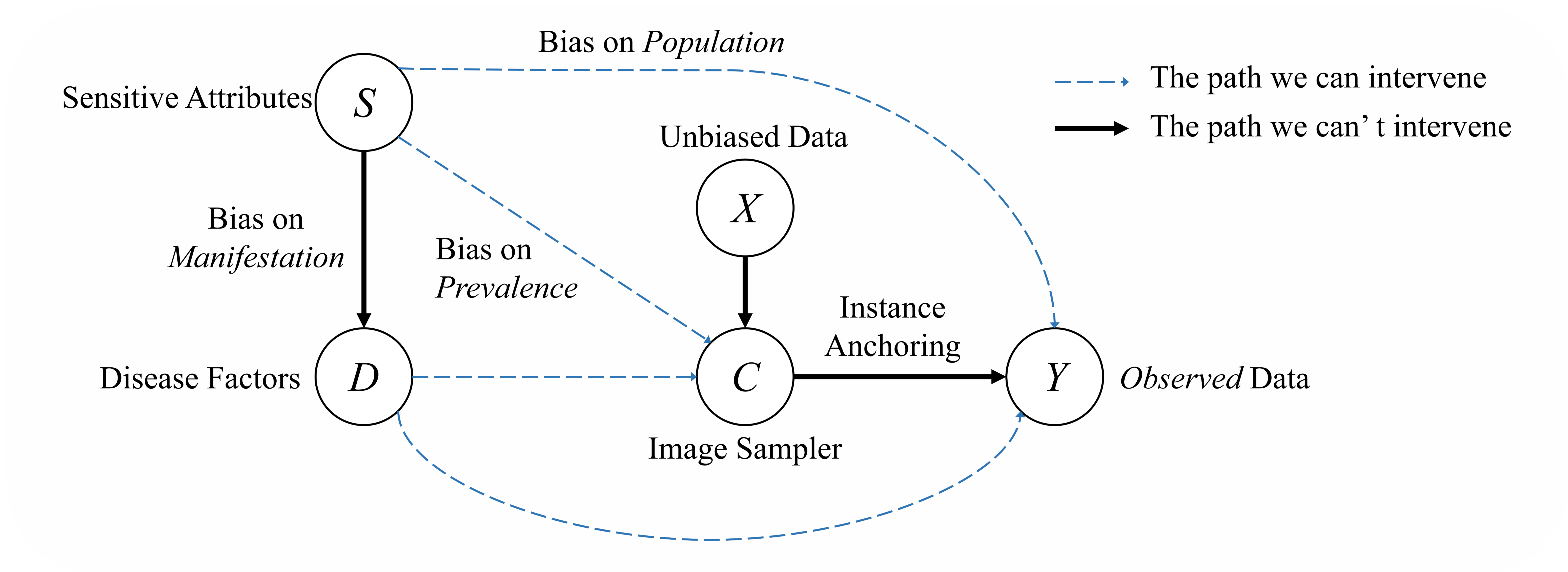}
 \caption{Causal graph of medical image formation. The graph illustrates how sensitive attributes ($S$), disease labels ($D$), and sampling mechanisms ($C$) influence the image ($Y$) through distinct causal pathways and biases.}
  \label{fig:framework}
\end{figure}

\paragraph{Setup and notation.}
As illustrated in Fig.~\ref{fig:framework}, our structural causal model (SCM) elucidates how subgroup unfairness emerges even in the absence of explicit access to $S$. Because the sensitive attribute fundamentally shapes the training distribution, we formalize this bias through four distinct causal pathways as follows:
\begin{equation}
S \rightarrow Y,\quad
S \rightarrow D \rightarrow Y,\quad
S \rightarrow C \rightarrow Y,\quad
S \rightarrow D \rightarrow C \rightarrow Y.
\end{equation}

We detail the underlying causal mechanisms for each pathway below:

\noindent\textbf{1. $S \rightarrow Y$: Population bias.}
This represents a direct causal influence where sensitive attributes induce group-specific appearance or acquisition shifts. Consequently, the interventional distribution $P(Y|do(S))$ allowing models to leverage these cues as shortcuts. This results in subgroup-specific error patterns even when the clinical label is invariant~\cite{gichoya2022race_recognition,lotter2024acquisition_race,geirhos2020shortcut_learning}.

\noindent\textbf{2. $S \rightarrow D \rightarrow Y$: Manifestation bias.}
Here, $S$ acts as a modifier of the disease's visual expression. Even with disease $D$ is held constant, the conditional distribution $P(Y|D,S)$ varies due to subgroup-linked differences in anatomical context or severity.
Models trained on dominant phenotypes fails to generalize to underrepresented manifestations, leading to systematic diagnostic gaps~\cite{seyyedkalantari2021underdiagnosis_bias,larrazabal2020gender_imbalance,daneshjou2022derm_ai_disparities,chen2023algorithmic_fairness}.

\noindent\textbf{3. $S \rightarrow C \rightarrow Y$: Prevalence bias.}
In this pathway, $S$ influences the selection mechanism $C$, which governs data availability. This causal chain shifts the effective support of the observed distribution, creating a disparity where $P(Y|S, C=1)$ is poorly sampled for minority groups. This imbalance biases generative and predictive models toward majority-group image statistics~\cite{irvin2019chexpert,johnson2019mimiccxr,glocker2023protected_characteristics_cxr,lotter2024acquisition_race}.

\noindent\textbf{4. $S \rightarrow D \rightarrow C \rightarrow Y$: Environment bias.}
This pathway represents intersectional selection bias. Formally, the selection mechanism $P(C | S, D)$ creates non-random missingness that varies across demographic-clinical "slices." Unlike marginal bias, this concentrates failures in specific intersectional strata, such as a specific disease manifestation unique to a vulnerable subgroup~\cite{seyyedkalantari2021underdiagnosis_bias}.

\begin{table}[!t]
\centering
\footnotesize
\caption{CXR examples for causal pathways by which sensitive factors $S$ can influence the observed image $Y$.}
\label{tab:graph_examples}
\setlength{\tabcolsep}{4pt}
\renewcommand{\arraystretch}{1.25}
\begin{tabularx}{\linewidth}{
    >{\centering\arraybackslash}p{0.16\linewidth}
    >{\raggedright\arraybackslash}X
}
\toprule
\textbf{Causal pathway} & \textbf{Phenomenon in CXR} \\
\midrule

$S \rightarrow Y$ &
Base physiological differences (e.g., bony patterns, global intensity) that affect the background of an image regardless of disease. \\
\addlinespace[2pt]

$S \rightarrow D \rightarrow Y$ &
Group-specific presentation of a disease (e.g., how "Pneumothorax" looks different in a smaller chest cavity vs. a larger one). \\
\addlinespace[2pt]

$S \rightarrow C \rightarrow Y$ &
Either because a group is sampled less often or has a different natural base rate of the disease (e.g., White 56.7\%, Asian $<11.0\%$). \\
\addlinespace[2pt]

$S \rightarrow D \rightarrow C \rightarrow Y$ &
The complex interaction where the likelihood of a patient being "captured" depends on both their demographic (e.g. Older patients are twice as likely to have respiratory issues like edema). \\
\bottomrule
\end{tabularx}
\end{table}

To ground these concepts, Table~\ref{tab:graph_examples} provides real-world examples of these biases in Chest X-ray (CXR) imaging. Our causal analysis implies simply increasing the volume of synthetic data is insufficient; rather, because distinct disparities originate from specific structural dependencies, they necessitate targeted intervention targets.
Accordingly, CIPHER explicitly models $S$, $D$, and $C$, via two complementary controls:
(i) Global $(S,D)$ Conditioning: Controls subgroup and disease composition to decouple $S$ from $D$, mitigating $S \to Y$ shortcuts and isolating manifestation shifts in $P(Y|D, S)$.
(ii) Instance-Anchored Counterfactuals: Employs local edits to generate "matched pairs" that bypass the selection mechanism $C$, populating sparse $(S, D)$ strata to counteract prevalence and environment biases.
This transforms augmentation from a black-box heuristic into a causal framework for addressing the structural sources of subgroup disparity.

\section{Methodology}
\subsection{Preliminaries: Latent Diffusion and Null-Text Inversion}
\label{sec:prelim}

\noindent\textbf{Latent diffusion.}
As shown in Fig.~\ref{fig:pipeline}, we adopt a latent diffusion model (LDM) that runs diffusion in a compact latent space \cite{ho2020ddpm,nichol2021improved_ddpm,rombach2022latent_diffusion}. At its core, the model relies on a text-conditioned U-Net $\epsilon_\theta(\cdot)$ to estimate the noise injected into the latent space. By conditioning on the text embedding $c$, the network denoises the representation to recover the target latent state at any given step $t$.

\noindent\textbf{Classifier-free guidance (CFG).}
We use classifier-free guidance:
\begin{equation}
\hat{\epsilon} = \epsilon_\theta(z_t,t,u) + s\big(\epsilon_\theta(z_t,t,c)-\epsilon_\theta(z_t,t,u)\big),
\end{equation}
where $u$ is the unconditional embedding and $s$ is the guidance scale \cite{ho2021classifier_free_guidance}.

\noindent\textbf{Null-text inversion for faithful, editable reconstruction.}
To realize our counterfactual editing function, $\mathcal{E}$, which executes a designated attribute transition (denoted by the operator $\rightarrow$), we employ null-text inversion~\cite{mokady2023null_text_inversion}.
This technique optimizes only the unconditional embeddings in CFG to improve image reconstruction while keeping $c$ fixed.
Given an input anchor image and its source prompt $c$ (capturing the orignal attributes ($d, s$)), we perform DDIM inversion to obtain an initial trajectory \cite{song2021ddim}, then optimize step-wise unconditional embeddings $\{u_t\}_{t=1}^T$ to minimize reconstruction error.
To apply a counterfactual transition such as $(d,s)\rightarrow(d,s')$ during the editing phase, we replace $c$ with an edited prompt $c'$ (e.g., modifying disease/metadata tokens).  By generating the new image with $c'$ while reusing the optimized $\{u_t\}_{t=1}^T$, the function $\mathcal{E}$ successfully modifies the targeted attributes while preserving the unobserved image structure, drastically reducing off-target drift.
We support global (unmasked) and localized (masked) edits under the same inversion trajectory (Fig.~\ref{fig:pipeline}) \cite{meng2022sdedit,hertz2022prompt_to_prompt,brooks2023instructpix2pix}.

\subsection{Causal Framework and Intervenable Pathways}
\label{sec:synth}

\begin{figure}[t]
  \centering
  \includegraphics[width=\linewidth]{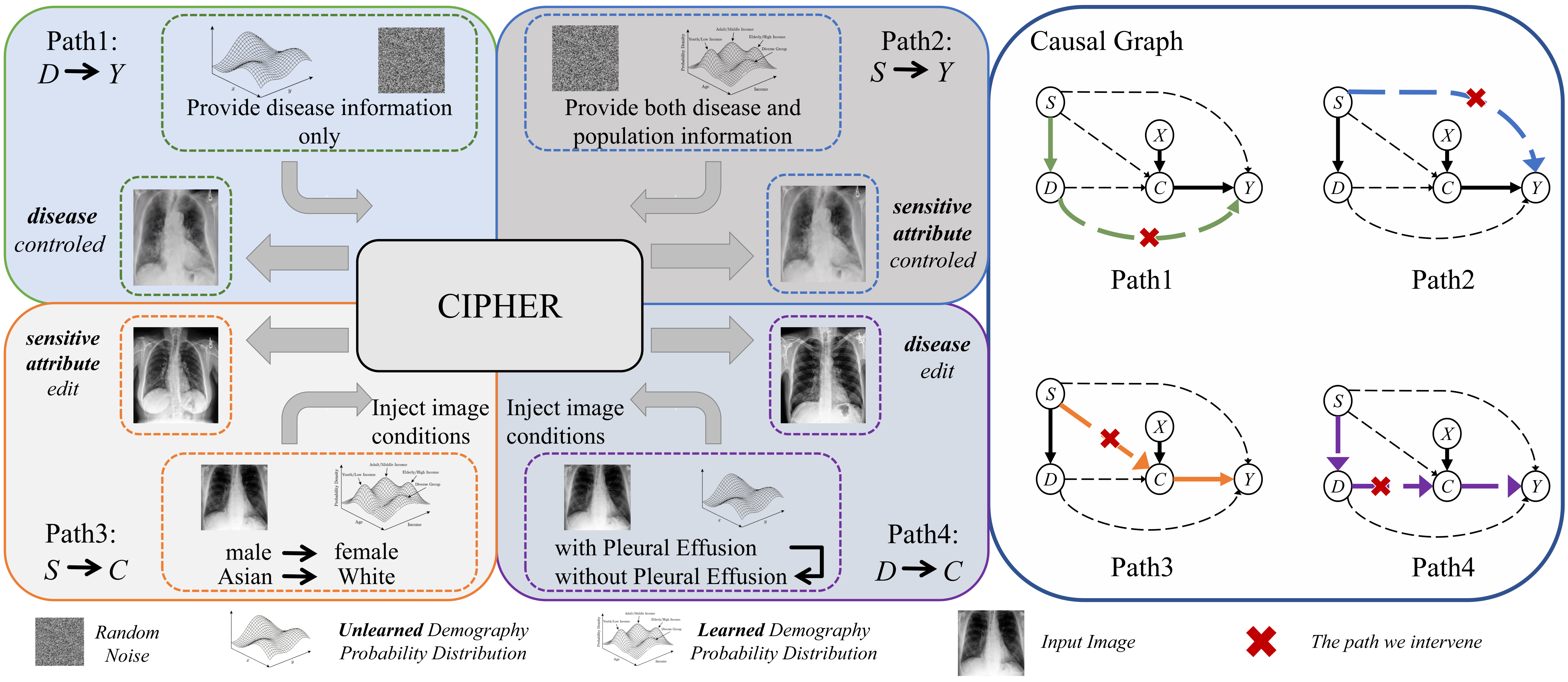}
  \caption{Overview of our causal counterfactual generation framework.}
  \label{fig:pipeline}
\end{figure}

\noindent\textbf{Causal Variables and the Generation Switch.}
We treat the textual representations of the sensitive and disease variables as intervenable—meaning we can inject, edit, block, or hold fixed their respective text embeddings, $c_S(\cdot)$ and $c_D(\cdot)$. 
Crucially, we do not intervene on intrinsic, dataset-implied constraints (such as the biased observational co-occurrence of $S$ and $D$) or the instance-anchoring structural noise when editing a real image.

\noindent\textbf{Four Complementary Generation and Editing Paths.}
Based on this framework, Fig.~\ref{fig:pipeline} instantiates into four parallel paths:
These pathways operate at two distinct rungs of the causal hierarchy, dictated by $C$: Paths 1 and 2 operate as global synthesis directly from noise, while Paths 3 and 4 perform instance-anchored counterfactual editing via diffusion inversion on real images.

\noindent\textbf{Path 1: Disease-only generation (control $D$, block $S$).}
When generating from noise, we sample from the observational distribution by omitting the sensitive tokens entirely:
\begin{equation}
y \sim p_\theta\!\left(Y \mid c_D(d')\right).
\end{equation}
By conditioning solely on $c_D(d')$, we block explicit textual guidance for $S$. While this increases the broader appearance coverage under a given disease $D$, the generative process marginalizes over the biased observational prior.

\noindent\textbf{Path 2: Sensitive-focused generation (control $S$ with $D$ held fixed).}
To perform an interventional cross-demographic synthesis, we apply a joint intervention on both variables by concatenating (i.e. $\oplus$) the conditions:
\begin{equation}
y \sim p_\theta\!\left(Y \mid c_D(d)\,\oplus\,c_S(s')\right).
\end{equation}
We hold the disease semantics $D{=}d$ fixed while systematically varying $S$ to reshape subgroup composition.  This forces the model to decouple the variables, allowing us to generate balanced datasets and actively probe subgroup-associated visual cues without altering the core clinical manifestation.

\noindent\textbf{Path 3: Instance-anchored sensitive counterfactual editing (edit $S$, keep $D$ fixed).} Transitioning to counterfactual editing, we intervene on $S$ for a specific observed anchor image $X=x$:
\begin{equation}
y \;=\; \mathcal{E}(x; (d,s)\rightarrow(d,s')).
\end{equation}
Using diffusion inversion to capture the unobserved background context of the real instance, we edit only $S$-related tokens while holding $D$ structurally unchanged. This yields strictly matched counterfactuals, effectively isolating the demographic features to audit the model's reliance on shortcut cues.

\noindent\textbf{Path 4: instance-anchored disease counterfactual editing (edit $D$, keep $S$ fixed).} Similarly, we compute the counterfactual manifestation of the disease on a fixed patient background:
\begin{equation}
y \;=\; \mathcal{E}(x; (d,s)\rightarrow(d',s)).
\end{equation}
We abduce the latent context of $x$ and selectively swap the disease condition $c_D(d)\!\rightarrow\!c_D(d')$ during the denoising trajectory while keeping $S$ strictly fixed. This ensures that the causal effect of the edit preserves the specific patient context and concentrates changes entirely on the physical evidence of the disease.


\section{Experiments and Results}
\noindent\textbf{Datasets.}
We use CheXpert Plus as the in-distribution (InD) dataset and MIMIC-CXR as the out-of-distribution (OOD) testbed \cite{irvin2019chexpert,johnson2019mimiccxr}.
For CXR, we apply subject-disjoint InD splits, and reserve MIMIC-CXR for OOD evaluation only.
We restrict images to the PA view, remove uncertain labels, and select 6 disease labels, yielding 14,138 InD training images and 8,347 OOD test images under the same criteria.
We leverage report text to guide diffusion training~\cite{irvin2019chexpert}. For dermoscopy, we use the public MILK10K dataset~\cite{tschandl2026milk10k}, which includes demographic metadata (age, sex, and skin-tone proxies).
We use all 10,480 images to fine-tune the diffusion model.

\noindent\textbf{Evaluation.}
We evaluate performance across age and race subgroups: ages 21--90 are binned into 7 ten-year groups (21--30, \dots, 81--90), and CXR race analysis considers the four most populated categories---Asian, White, Black, and Hispanic.In addition, on the dermoscopy dataset, we further stratify test cases by the \textit{MILK10K skin-tone level} labels for subgroup evaluation.
For downstream diagnosis, we train a DenseNet121 \cite{huang2017densenet} and report test-set mAUC.
Fairness is assessed via subgroup AUROC per demographic attribute. We report the maximum AUROC gap across categories (Age/Sex/Race/Skin) and its relative reduction compared the Path1 (i.e. disease-conditioned synthesis) baseline, a widely adopted strategy for enhancing model performance.

\subsection{Fine-Tune Diffusion Model}
\label{sec:finetune_diffusion}

We fine-tune \textbf{Stable Diffusion v1-4} on each modality to adapt the generator to the medical imaging domain\cite{ho2020ddpm,nichol2021improved_ddpm,rombach2022latent_diffusion}. 
For all experiments, we run 30,000 optimization steps with a base learning rate of $1\times10^{-4}$ and a cosine annealing schedule. Unless specified, we maintain a 1:1 real-to-synthetic ratio. Evaluating ratios from 1:1.5 to 4:1 confirms that 1:1 yields the optimal accuracy–fairness trade-off.
We freeze all components except the U-Net and the text encoder. 

For CXR, we fine-tune using cleaned medical report--image text pairs, where textual condition is derived from the processed radiology reports.
For dermoscopy, we fine-tune using disease label--image text pairs, i.e., prompts constructed from diagnostic labels.
In both modalities, we additionally inject metadata tokens (e.g., age/sex/race or skin-tone when available) into the textual condition to enable demographic-aware generation and editing in subsequent paths.

\subsection{Ablation Experiments}
\label{sec:ablation}
\begin{figure}[t]
  \centering
  \includegraphics[width=\linewidth]{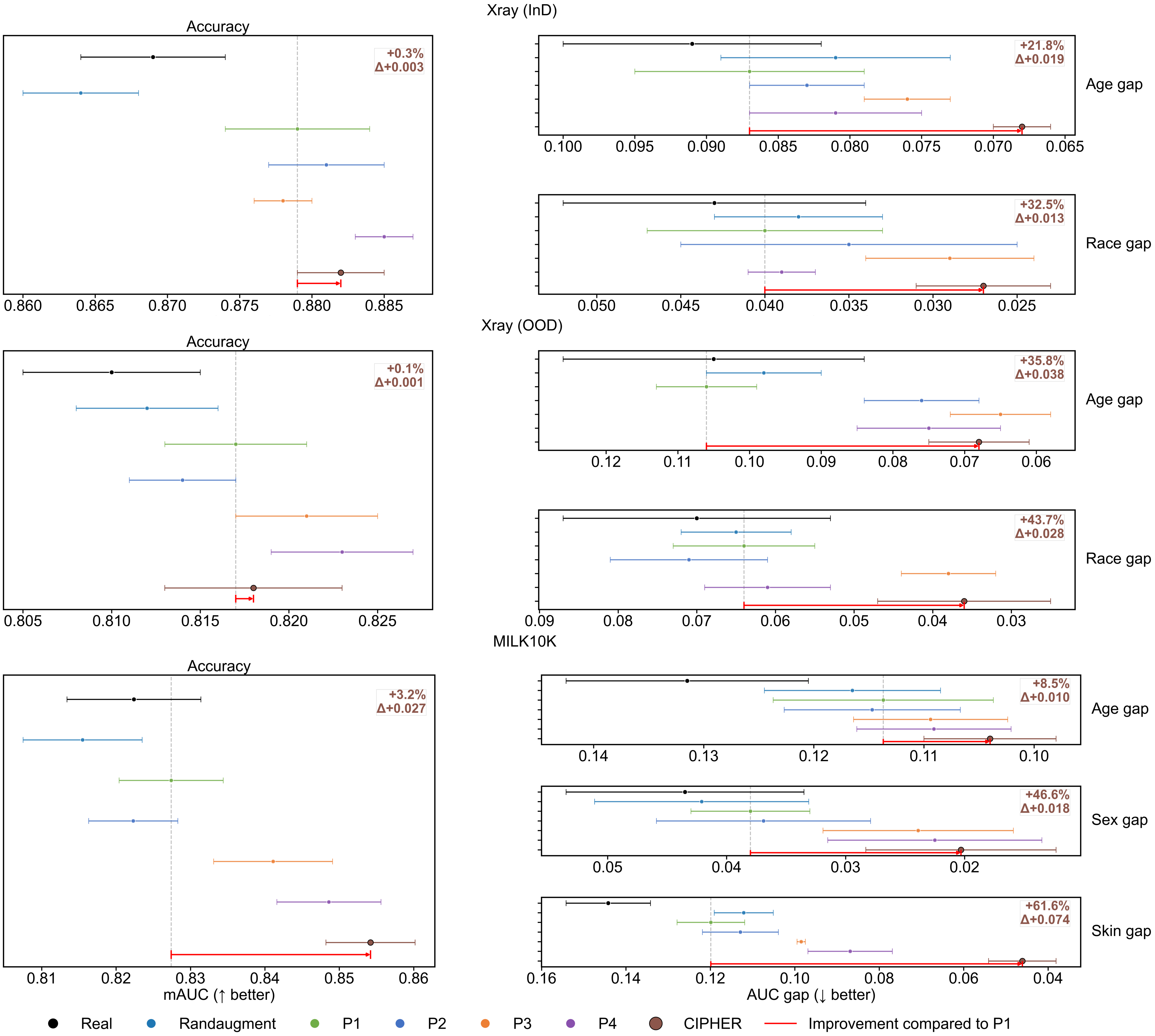}
 \caption{Results on chest radiology and dermoscopy datasets: average AUC vs. fairness (AUC) gap across different paths for race/age (radiology, in-distribution \& OOD) and sex/age/skin tone (dermoscopy).}
  \label{fig:result}
\end{figure}

To evaluate CIPHER for downstream diagnosis and disentangle the effects of different paths, we train classifiers under seven data settings: Real, a conventional augmentation baseline RandAugment, P1--P4 (Real augmented with synthetic samples generated by a \emph{single} path), and CIPHER (our full method).
In Fig.~\ref{fig:result}, each marker shows the mean with error bars across runs; moving right indicates a better performance and smaller gap (better fairness).

Upper Fig.~\ref{fig:result} illustrates CXR performance and fairness for Age and Race under InD (CheXpert Plus) and OOD (MIMIC) shifts.
\textbf{InD:} single-path variants can improve mAUC (e.g., P4 is highest), but gap reductions are limited. CIPHER yields the best overall trade-off, improving mAUC by 0.003 while reducing the Age gap by 0.019 ({21.8\%}) and the Race gap by 0.013 ({32.5\%}) relative to P1.
\textbf{OOD:} RandAugment and single-path variants are less consistent under shift, often improving mAUC without reliably reducing gaps. CIPHER remains the most robust setting, improving mAUC by 0.001 and reducing the Age gap by 0.038 ({35.8\%}) and the Race gap by 0.028 ({43.7\%}).

Lower Fig.~\ref{fig:result} reports MILK10K performance and fairness for Age, {Sex}, and {Skin}. Single-path variants provide partial gains, whereas CIPHER consistently dominates, achieving the highest mAUC ({0.027}) and the largest gap reductions: {Age} {0.010} ({8.5\%}), {Sex} {0.023} ({46.6\%}), and {Skin} {0.074} ({61.6\%}).

We further compared CIPHER with MixUp, CutMix, JTT, and SELF across three datasets. CIPHER consistently yields the smallest fairness gaps across age, race, gender, and skin tone in both in-distribution and OOD settings, validating the advantage of our causality-driven framework over heuristic augmentation and debiasing methods.

\section{Conclusion}
We introduce CIPHER, a diffusion-based framework that optimizes the accuracy--fairness trade-off by leveraging a structural causal perspective of medical image formation.
By organizing control around $(S,D,C,Y)$ and intervening on controllable edges, CIPHER provides a unified approach to fairness-aware data augmentation. 
We show that synthesizing complementary causal pathways yields the most robust trade-off, simultaneously enhancing mean AUROC and narrowing subgroup disparities under distribution shift.
Future research could explore the integration of more sophisticated generative architectures or the development of a fully unified synthesis pipeline based on the presented SCM.

\vspace{1em}

\noindent\textbf{Disclosure of Interests.} The authors have no competing interests to declare that
are relevant to the content of this article.
\bibliographystyle{splncs04}
\bibliography{bibliography}

\end{document}